\documentclass[11pt]{article}
\usepackage{acl2016}
\usepackage{times}
\usepackage{latexsym}
\usepackage{amsmath,amssymb} 
\usepackage{graphicx,color,epsfig}
\usepackage{soul,url,tabulary}
\usepackage{cleveref}
\crefname{section}{§}{§§}
\Crefname{section}{§}{§§}

\aclfinalcopy 
\def\aclpaperid{41} 

\usepackage{color}
\usepackage{amsmath}
\usepackage{amssymb}
\usepackage{multirow, varwidth}
\usepackage{stfloats}
\usepackage{verbatim}
\makeatletter
\newif\if@restonecol
\makeatother

\usepackage[ruled,vlined,lined,boxed,linesnumbered]{algorithm2e}
\let\oldnl\nl
\newcommand{\nonl}{\renewcommand{\nl}{\let\nl\oldnl}}
\usepackage{xr}
\usepackage[amsmath,thmmarks]{ntheorem}

\theorembodyfont{\normalfont \it}
\theoremseparator{.}

\theoremstyle{nonumberplain}

\newcommand{\bm}[1]{\boldsymbol{#1}}

\newcommand{\argmin}{\operatornamewithlimits{argmin}}

\DeclareRobustCommand\onedot{\futurelet\@let@token\@onedot}
\def\onedot{. }
 
\def\ie{\emph{i.e}\onedot}

\newcolumntype{K}[1]{>{\centering\arraybackslash}p{#1}}

\newcolumntype{P}[1]{>{\centering\arraybackslash}p{#1}}

\title{Learning Concept Taxonomies from Multi-modal Data}
\author{Hao Zhang$^1$, Zhiting Hu$^1$, Yuntian Deng$^1$, Mrinmaya Sachan$^1$, \\
\bf{Zhicheng Yan$^2$, Eric P. Xing$^1$}\\
	    $^1$Carnegie Mellon University, $^2$UIUC\\
	    {\tt \{hao,zhitingh,yuntiand,mrinmays,epxing\}@cs.cmu.edu}}
\date{}
\begin{document}
\maketitle
\begin{abstract}
We study the problem of automatically building hypernym taxonomies from textual and visual data. Previous works in taxonomy induction generally ignore the increasingly prominent visual data, which encode important perceptual semantics. Instead, we propose a probabilistic model for taxonomy induction by jointly leveraging text and images. To avoid hand-crafted feature engineering, we design end-to-end features based on distributed representations of images and words. The model is discriminatively trained given a small set of existing ontologies and is capable of building full taxonomies from scratch for a collection of unseen conceptual label items with associated images. We evaluate our model and features on the WordNet hierarchies, where our system outperforms previous approaches by a large gap. 
\end{abstract}

\section{Introduction}
Human knowledge is naturally organized as semantic hierarchies. For example, in WordNet \cite{miller1995wordnet}, specific concepts are categorized and assigned to more general ones, leading to a semantic hierarchical structure (a.k.a taxonomy). A variety of NLP tasks, such as question answering~\cite{harabagiu2003open}, document clustering~\cite{hotho2002ontology} and text generation \cite{biran2013classifying} can benefit from the conceptual relationship present in these hierarchies.

\begin{figure}[t]
\small \centering \hfill \hspace{-20pt}
\includegraphics[width=8cm]{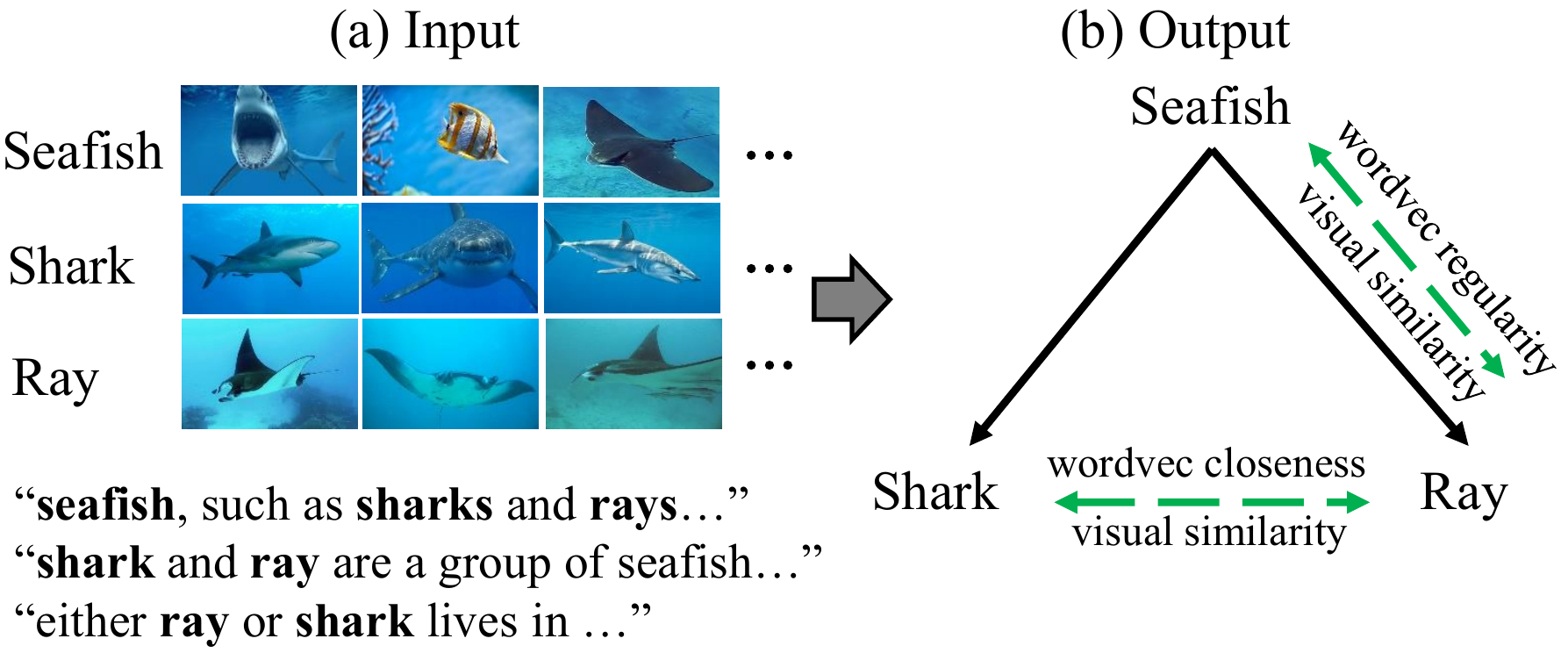}
\vspace{-10pt}
\caption{An overview of our system. (a) Input: a collection of label items, represented by text and images; (b) Output: we build a taxonomy from scratch by extracting features based on distributed representations of text and images.}
\vspace{-10pt}
\label{fig:teaser}
\end{figure}

Traditional methods of manually constructing taxonomies by experts (e.g. WordNet) and interest communities (e.g. Wikipedia) are either knowledge or time intensive, and the results have limited coverage. Therefore, automatic induction of taxonomies is drawing increasing attention in both NLP and computer vision. On one hand, a number of methods have been developed to build hierarchies based on lexical patterns in text~\cite{yang2009metric,snow2006semantic,kozareva2010semi,navigli2011graph,fu2014learning,bansalstructured,tuan2015incorporating}. These works generally ignore the rich visual data which encode important perceptual semantics~\cite{bruni2014multimodal} and have proven to be complementary to linguistic information and helpful for many tasks~\cite{silberer2014learning,kiela2014learning,zhang2015marketcompetition,chen2013neil}. On the other hand, researchers have built visual hierarchies by utilizing only visual features~\cite{griffin2008learning,yan2015hd,sivic2008unsupervised}. The resulting hierarchies are limited in interpretability and usability for knowledge transfer.

Hence, we propose to combine both visual and textual knowledge to automatically build taxonomies.
We induce {\it is-a} taxonomies by 
supervised learning from existing entity ontologies where each concept category (entity) is associated with images, either from existing dataset (e.g. ImageNet \cite{deng2009imagenet}) or retrieved from the web using search engines, as illustrated in Fig \ref{fig:teaser}. Such a scenario is realistic and can be extended to a variety of tasks; for example, in knowledge base construction \cite{chen2013neil}, text and image collections are readily available but label relations among categories are to be uncovered. In large-scale object recognition, automatically learning relations between labels can be quite useful \cite{deng2014large,zhao2011large}.

Both textual and visual information provide important cues for taxonomy induction. Fig \ref{fig:teaser} illustrates this via an example. The parent category \emph{seafish} and its two child categories \emph{shark} and \emph{ray} are closely related as: (1) there is a hypernym-hyponym (\emph{is-a}) relation between the words ``seafish'' and ``shark''/``ray'' through text descriptions like ``...seafish, such as shark and ray...'', ``...shark and ray are a group of seafish...''; (2) images of the close neighbors, e.g., \emph{shark} and \emph{ray} are usually visually similar and images of the child, e.g. \emph{shark/ray} are similar to a subset of images of \emph{seafish}. To effectively capture these patterns, in contrast to previous works that rely on various hand-crafted features \cite{chen2013neil,bansalstructured}, we extract features by leveraging the {\it distributed representations} that embed images \cite{simonyan2014very} and words \cite{mikolov2013distributed} as compact vectors, based on which the semantic closeness is directly measured in vector space.
Further, we develop a probabilistic framework that integrates the rich multi-modal features to induce ``is-a'' relations between categories, encouraging {\it local semantic consistency} that each category should be visually and textually close to its parent and siblings.


In summary, this paper has the following contributions: (1) We propose a novel probabilistic Bayesian model (Section\cref{sec:model}) for taxonomy induction by jointly leveraging textual and visual data. The model is discriminatively trained and can be directly applied to build a taxonomy from scratch for a collection of semantic labels. (2) We design novel features (Section\cref{sec:feat}) based on general-purpose distributed representations of text and images to capture both textual and visual relations between labels. (3) We evaluate our model and features on the ImageNet hierarchies with two different taxonomy induction tasks (Section\cref{sec:experiments}). We achieve superior performance on both tasks and improve the $F_1$ score by 2x in the \emph{taxonomy construction} task, compared to previous approaches.
Extensive comparisons demonstrate the effectiveness of integrating visual features with language features for taxonomy induction.
We also provide qualitative analysis on our features, the learned model, and the taxonomies induced to provide further insights (Section~\ref{sec:exp-qualitative}).

\section{Related Work}
Many approaches have been recently developed that build hierarchies purely by identifying either lexical patterns or statistical features in text corpora \cite{yang2009metric,snow2006semantic,kozareva2010semi,navigli2011graph,zhu2013topic,fu2014learning,bansalstructured,tuan2014taxonomy,tuan2015incorporating,kiela2015exploiting}. The approaches in \newcite{yang2009metric} and \newcite{snow2006semantic} assume a starting incomplete hierarchy and try to extend it by inserting new terms. \newcite{kozareva2010semi} and \newcite{navigli2011graph} first find leaf nodes and then use lexical patterns to find intermediate terms and all the attested hypernymy links between them. In \cite{tuan2014taxonomy}, syntactic contextual similarity is exploited to construct the taxonomy, while \newcite{tuan2015incorporating} go one step further to consider trustiness and collective synonym/contrastive evidence. Different from them, our model is discriminatively trained with multi-modal data.
The works of \newcite{fu2014learning} and \newcite{bansalstructured} use similar language-based features as ours. Specifically, in \cite{fu2014learning}, linguistic regularities between pretrained word vectors \cite{mikolov2013distributed} are modeled as projection mappings. The trained projection matrix is then used to induce pairwise hypernym-hyponym relations between words. 
Our features are partially motivated by \newcite{fu2014learning}, but we jointly leverage both textual and visual information.
In \newcite{kiela2015exploiting}, both textual and visual evidences are exploited to detect pairwise lexical entailments. Our work is significantly different as our model is optimized over the whole taxonomy space rather than considering only word pairs separately.
In \cite{bansalstructured}, a structural learning model is developed to induce a globally optimal hierarchy.
Compared with this work, we exploit much richer features from both text and images, and leverage distributed representations instead of hand-crafted features.

Several approaches \cite{griffin2008learning,bart2008unsupervised,marszalek2008constructing} have also been proposed to construct visual hierarchies from image collections.
In \cite{bart2008unsupervised}, a nonparametric Bayesian model is developed to group images based on low-level features. In \cite{griffin2008learning} and \cite{marszalek2008constructing}, a visual taxonomy is built to accelerate image categorization. In \cite{chen2013neil}, only binary object-object relations are extracted using co-detection matrices.
Our work differs from all of these as we integrate textual with visual information to construct taxonomies.

Also of note are several works that integrate text and images as evidence for knowledge base autocompletion \cite{bordes2011learning} and zero-shot recognition \cite{gan2015exploring,gan2016recognizing,socher2013zero}. Our work is different because our task is to accurately construct multi-level hyponym-hypernym hierarchies from a set of (seen or unseen) categories.

\section{Taxonomy Induction Model}
\label{sec:model}

Our model is motivated by the key observation that in a semantically meaningful taxonomy, a category tends to be closely related to its children as well as its siblings. For instance, there exists a hypernym-hyponym relation between the name of category {\it shark} and that of its parent {\it seafish}. Besides, images of {\it shark} tend to be visually similar to those of {\it ray}, both of which are seafishes. Our model is thus designed to encourage such local semantic consistency; and by jointly considering all categories in the inference, a globally optimal structure is achieved. A key advantage of the model is that we incorporate both visual and textual features induced from distributed representations of images and text (Section~\ref{sec:feat}). These features capture the rich underlying semantics and facilitate taxonomy induction. We further distinguish the relative importance of visual and textual features that could vary in different layers of a taxonomy. Intuitively, visual features would be increasingly indicative in the deeper layers, as sub-categories under the same category of specific objects tend to be visually similar. In contrast, textual features would be more important when inducing hierarchical relations between the categories of general concepts (i.e. in the near-root layers) where visual characteristics are not necessarily similar.

\subsection{The Problem}
Assume a set of $N$ categories $\bm{x}=\{x_1, x_2, \dots, x_N\}$, where each category $x_n$ consists of a text term $t_{n}$ as its name, as well as a set of images $\bm{i}_{n} = \{i_1, i_2, \dots\}$. Our goal is to construct a taxonomy tree $T$ over these categories\footnote{We assume $T$ to be a tree. Most existing taxonomies are modeled as trees \cite{bansalstructured}, since a tree helps simplify the construction and ensures that the learned taxonomy is interpretable. With minor modifications, our model also works on non-tree structures.}, such that categories of specific object types (e.g. shark) are grouped and assigned to those of general concepts (e.g. seafish). As the categories in $\bm{x}$ may be from multiple disjoint taxonomy trees, we add a {\it pseudo} category $x_0$ as the hyper-root so that the optimal taxonomy is ensured to be a single tree. Let $z_{n}\in\{1,\dots, N\}$ be the index of the parent of category $x_n$, i.e.  $x_{z_{n}}$ is the hypernymic category of $x_{n}$. Thus the problem of inducing a taxonomy structure is equivalent to inferring the conditional distribution $p(\bm{z} | \bm{x})$ over the set of (latent) indices $\bm{z} = \{ z_1, \dots, z_n \}$, based on the images and text.

\subsection{Model}
We formulate the distribution $p(\bm{z} | \bm{x})$ through a model which leverages rich multi-modal features.
Specifically, let $\bm{c}_{n}$ be the set of child nodes of category $x_n$ in a taxonomy encoded by $\bm{z}$.
Our model is defined as
\vspace{-10pt}
\begin{equation}
\small
p_{w}(\bm{z}, \bm{\pi} | \bm{x}, \bm{\alpha}) \propto
p(\bm{\pi} | \bm{\alpha}) \prod_{n=1}^{N} \prod_{x_{n'}\in \bm{c}_{n}} \pi_{n} g_{w}(x_{n}, x_{n'}, \bm{c}_n \backslash x_{n'})
\label{eq:model-pi}
\vspace{-10pt}
\end{equation}
where $g_w(x_{n}, x_{n'}, \bm{c}_n \backslash x_{n'})$, defined as
\begin{equation*}
\small
g_{w}(x_n, x_{n'}, \bm{c}_n\backslash x_{n'}) = \exp \{ \bm{w}^\top_{d(x_{n'})} \bm{f}_{n, n', \bm{c}_n\backslash x_{n'}} \},
\label{eq:g}
\vspace{-5pt}
\end{equation*}
measures the semantic consistency between category $x_{n'}$, its parent $x_n$ as well as its siblings indexed by $\bm{c}_n \backslash x_{n'}$. The function $g_w(\cdot)$ is loglinear with respect to $\bm{f}_{n, n', \bm{c}_n\backslash x_{n'}}$, which is the feature vector defined over the set of relevant categories $( x_n, x_{n'}, \bm{c}_n\backslash x_{n'} )$, with $\bm{c}_{n} \backslash x_{n'}$ being the set of child categories excluding $x_{n'}$ (Section \ref{sec:feat}). The simple exponential formulation can effectively encourage close relations among nearby categories in the induced taxonomy.
The function has combination weights $\bm{w} = \{\bm{w}_1, \dots, \bm{w}_{L}\}$, where $L$ is the maximum depth of the taxonomy, to capture the importance of different features, and the function $d(x_{n'})$ to return the depth of $x_{n'}$ in the current taxonomy. Each layer $l\ (1\leq l \leq L)$ of the taxonomy has a specific $\bm{w}_{l}$ thereby allowing varying weights of the same features in different layers. The parameters are learned in a {\it supervised} manner.
In eq~\ref{eq:model-pi}, we also introduce a weight $\pi_n$ for each node $x_n$, in order to capture the varying popularity of different categories (in terms of being a parent category). For example, some categories like {\it plant} can have a large number of sub-categories, while others such as {\it stone} have less. We model $\bm{\pi}$ as a multinomial distribution with Dirichlet prior $\bm{\alpha} = (\alpha_1, \dots, \alpha_N)$ to encode any prior knowledge of the category popularity\footnote{$\bm{\alpha}$ could be estimated using training data.}; and the conjugacy allows us to marginalize out $\bm{\pi}$ analytically to get
\vspace{-10pt}
\begin{equation}
\small
\begin{aligned}
p_{w}(\bm{z} | \bm{x}, \bm{\alpha}) & \propto \hspace{-4pt} \int \hspace{-2pt} p(\bm{\pi} | \bm{\alpha}) \prod_{n=1}^{N} \hspace{-2pt} \prod_{ x_{n'} \in \bm{c}_{n}} \hspace{-5pt} \pi_{n} g_{w}(x_{n}, x_{n'}, \bm{c}_n \backslash x_{n'}) d{\bm{\pi}} \\
&\propto \prod_{n} \Gamma(q_{n}+\alpha_n) \prod_{ x_{n'}\in \bm{c}_{n}} g_{w}(x_{n}, x_{n'}, \bm{c}_n \backslash x_{n'})
\end{aligned}
\label{eq:model}
\vspace{-5pt}
\end{equation}
where $q_n$ is the number of children of category $x_n$.

Next, we describe our approach to infer the expectation for each $z_n$, and based on that select a particular taxonomy structure for the category nodes $\bm{x}$. As $\bm{z}$ is constrained to be a tree (i.e. cycle without loops), we include with eq~\ref{eq:model}, an indicator factor $\bm{1}(\bm{z})$ that takes $1$ if $\bm{z}$ corresponds a tree and $0$ otherwise. We modify the inference algorithm appropriately to incorporate this constraint.

\noindent \textbf{Inference.}
Exact inference is computationally intractable due to the normalization constant of eq~\ref{eq:model}. We therefore use Gibbs Sampling, a procedure for approximate inference.
Here we present the sampling formula for each $z_n$ directly, and defer the details to the supplementary material. The sampling procedure is highly efficient because the normalization term and the factors that are irrelevant to $z_n$ are cancelled out. The formula is
\begin{equation}
\small
\begin{aligned}
p(z_{n} = & m | \bm{z}\backslash z_{n}, \cdot) \propto \bm{1}(z_{n} = m, \bm{z}\backslash z_{n}) \cdot \left( q_{m}^{-n} + \alpha_{m} \right) \cdot \\
& \frac{\prod_{x_{n'} \in \bm{c}_{m}\cup \{x_{n}\}} g_{w}(x_m, x_{n'}, \bm{c}_{m}\cup\{x_n\})}{\prod_{x_{n'} \in \bm{c}_{m}\backslash x_{n}} g_{w}(x_m, x_{n'}, \bm{c}_{m}\backslash x_n)},
\end{aligned}
\label{eq:gs}
\end{equation}
where $q_{m}$ is the number of children of category $m$; the superscript $-n$ denotes the number excluding $x_n$.
Examining the validity of the taxonomy structure (i.e. the tree indicator) in each sampling step can be computationally prohibitive. To handle this, we restrict the candidate value of $z_{n}$ in eq~\ref{eq:gs}, ensuring that the new $z_{n}$ is always a tree. Specifically, given a tree $T$, we define a {\it structure operation} as the procedure of detaching one node $x_n$ in $T$ from its parent and appending it to another node $x_{m}$ which is not a descendant of $x_n$. 

\begin{proposition*}
(1) Applying a structure operation on a tree $T$ will result in a structure that is still a tree.
(2) Any tree structure over the node set $\bm{x}$ that has the same root node with tree $T$ can be achieved by applying structure operation on $T$ a finite number of times.
\end{proposition*}

The proof is straightforward and we omit it due to space limitations. We also add a pseudo node $x_0$ as the fixed root of the taxonomy. Hence by initializing a tree-structured state rooted at $x_0$ and restricting each updating step as a structure operation, our sampling procedure is able to explore the whole valid tree space.

\noindent \textbf{Output taxonomy selection.}
To apply the model to discover the underlying taxonomy from a given set of categories, we first obtain the marginals of $\bm{z}$ by averaging over the samples generated through eq~\ref{eq:gs}, then output the optimal taxonomy $\bm{z}^{*}$ by finding the maximum spanning tree (MST) using the Chu-Liu-Edmonds algorithm  \cite{chu1965shortest,bansalstructured}.

\noindent \textbf{Training.}
We need to learn the model parameters $\bm{w}_{l}$ of each layer $l$, which capture the relative importance of different features. The model is trained using the EM algorithm.
Let $\ell(x_{n})$ be the depth (layer) of category $x_n$; and $\tilde{\bm{z}}$ (siblings $\tilde{\bm{c}}_n$) denote the gold structure in training data. Our training algorithm updates $\bm{w}$ through maximum likelihood estimation, wherein the gradient of $\bm{w}_{l}$ is (see the supplementary materials for details):
\begin{equation*}
\small
\begin{split}
\delta \bm{w}_{l} = \hspace{-8pt} \sum_{n:\ell(x_n) = l} \hspace{-8pt} \left\{ \bm{f}(x_{\tilde{z}_n}, x_n, \tilde{\bm{c}}_n\backslash x_n) \hspace{-2pt} - \hspace{-2pt} \mathbb{E}_{p}[ \bm{f}(x_{z_n}, x_n, \bm{c}_n\backslash x_n)] \right\},
\end{split}
\end{equation*}
which is the net difference between gold feature vectors and expected  feature vectors as per the model. The expectation is approximated by collecting samples using the sampler described above and averaging them.

\section{Features}
\label{sec:feat}
In this section, we describe the feature vector $\bm{f}$ used in our model, and defer more details in the supplementary material. Compared to previous taxonomy induction works which rely purely on linguistic information, we exploit both perceptual and textual features to capture the rich spectrum of semantics encoded in images and text. Moreover, we leverage the {\it distributed representations} of images and words to construct compact and effective features. Specifically, each image $i$ is represented as an embedding vector $\bm{v}_{i} \in \mathbb{R}^{a}$ extracted by deep convolutional neural networks. Such image representation has been successfully applied in various vision tasks. On the other hand, the category name $t$ is represented by its word embedding $\bm{v}_{t} \in \mathbb{R}^{b}$, a low-dimensional dense vector induced by the Skip-gram model \cite{mikolov2013distributed} which is widely used in diverse NLP applications too.
Then we design $\bm{f}(x_{n}, x_{n'}, \bm{c}_{n}\backslash x_{n'})$ based on the above image and text representations. The feature vector $\bm{f}$ is used to measure the local semantic consistency between category $x_{n'}$ and its parent category $x_{n}$ as well as its siblings $\bm{c}_{n}\backslash x_{n'}$.


\subsection{Image Features}
\noindent \textbf{Sibling similarity}.
As mentioned above, close neighbors in a taxonomy tend to be visually similar, indicating that the embedding of images of sibling categories should be close to each other in the vector space $\mathbb{R}^{a}$.
For a category $x_n$ and its image set $\bm{i}_n$, we fit a Gaussian distribution $\mathcal{N}(\overline{\bm{v}}_{\bm{i}_n}, \Sigma_{n})$ to the image vectors, where $\overline{\bm{v}}_{\bm{i}_n} \in \mathbb{R}^{a}$ is the mean vector and $\Sigma_{n} \in \mathbb{R}^{a\times a}$ is the covariance matrix. For a sibling category $x_m$ of $x_n$, we define the visual similarity between $x_n$ and $x_m$ as
\begin{equation*}
\small
\begin{aligned}
vissim(x_n, x_m) \hspace{-2pt} = \hspace{-2pt} [\mathcal{N}(\overline{\bm{v}}_{\bm{i}_m}; \overline{\bm{v}}_{\bm{i}_n}, \Sigma_{n})\hspace{-2pt} + \hspace{-2pt} \mathcal{N}(\overline{\bm{v}}_{\bm{i}_n}; \overline{\bm{v}}_{\bm{i}_m}, \Sigma_{m})]/2
\end{aligned}
\end{equation*}
which is the average probability of the mean image vector of one category under the Gaussian distribution of the other. This takes into account not only the distance between the mean images, but also the closeness of the images of each category.
Accordingly, we compute the visual similarity between $x_{n'}$ and the set $\bm{c}_{n}\backslash x_{n'}$ by averaging:
\begin{equation*}
\small
\begin{aligned}
vissim(x_{n'}, \bm{c}_{n}\backslash x_{n'}) = \frac{\sum_{x_m \in \bm{c}_{n}\backslash x_{n'}} vissim(x_{n'}, x_m)}{|\bm{c}_{n}|-1}.
\end{aligned}
\end{equation*}
We then bin the values of $vissim(x_{n'}, \bm{c}_{n}\backslash x_{n'})$ and represent it as an one-hot vector, which constitutes $\bm{f}$ as a component named as \emph{siblings image-image relation feature} (denoted as \emph{S-V1}\footnote{S: sibling, PC: parent-child, V: visual, T: textual. }).

\noindent\textbf{Parent prediction}.
Similar to feature S-V1, we also create the similarity feature between the image vectors of the parent and child, to measure their visual similarity. However, the parent node is usually a more general concept than the child, and it usually consists of images that are not necessarily similar to its child. Intuitively, by narrowing the set of images to those that are most similar to its child improves the feature. Therefore, different from S-V1, when estimating the Gaussian distribution of the parent node, we only use the top $K$ images with highest probabilities under the Gaussian distribution of the child node. We empirically show in section \ref{sec:exp-qualitative} that choosing an appropriate $K$ consistently boosts the performance. We name this feature as \emph{parent-child image-image relation feature} (denoted as \emph{PC-V1}).

Further, inspired by the linguistic regularities of word embedding, i.e. the hypernym-hyponym relationship between words can be approximated by a linear projection operator between word vectors \cite{mikolov2013distributed,fu2014learning}, we design a similar strategy to \cite{fu2014learning} between images and words so that the parent can be ``predicted'' given the image embedding of its child category and the projection matrix.
Specifically, let $(x_{n}, x_{n'})$ be a parent-child pair in the training data, we learn a projection matrix $\bm{\Phi}$ which minimizes the distance between $\bm{\Phi}\overline{\bm{v}}_{\bm{i}_{n'}}$ (i.e. the projected mean image vector $\overline{\bm{v}}_{\bm{i}_{n'}}$ of the child) and $\bm{v}_{t_n}$ (i.e. the word embedding of the parent):
\begin{equation*}
\small
\begin{split}
\bm{\Phi}^{*} = \argmin_{\bm{\Phi}} \frac{1}{N}\sum_{n}\| \bm{\Phi}\overline{\bm{v}}_{\bm{i}_{n'}} - \bm{v}_{t_n}\|_{2}^{2} + \lambda\|\bm{\Phi}\|_{1},
\end{split}
\end{equation*}
where $N$ is the number of parent-child pairs in the training data.
Once the projection matrix has been learned, the similarity between a child node $x_{n'}$ and its parent $x_{n}$ is computed as $\|\bm{\Phi}\overline{\bm{v}}_{\bm{i}_{n'}} - \bm{v}_{t_{n}}\|$, and we also create an one-hot vector by binning the feature value. We call this feature as \emph{parent-child image-word relation feature} (\emph{PC-V2}).

\subsection{Word Features}
We briefly introduce the text features employed. More details about the text feature extraction could be found in the supplementary material.

\noindent\textbf{Word embedding features}.d PC-V1, We induce features using word vectors to measure both sibling-sibling and parent-child closeness in text domain \cite{fu2014learning}. One exception is that, as each category has only one word, the sibling similarity is computed as the cosine distance between two word vectors (instead of mean vectors).
This will produce another two parts of features, \emph{parent-child word-word relation feature} (\emph{PC-T1}) and \emph{siblings word-word relation feature} (\emph{S-T1}).

\noindent \textbf{Word surface features}.
In addition to the embedding-based features, we further leverage lexical features based on the surface forms of child/parent category names. Specifically, we employ the \emph{Capitalization}, \emph{Ends with}, \emph{Contains}, \emph{Suffix match}, \emph{LCS} and \emph{Length different} features, which are commonly used in previous works in taxonomy induction \cite{yang2009metric,bansalstructured}.

%
%



\section{Experiments}\label{sec:experiments}
We first disclose our implementation details in section \ref{sec:exp-setting} and the supplementary material for better reproducibility. We then compare our model with previous state-of-the-art methods \cite{fu2014learning,bansalstructured} with two taxonomy induction tasks. Finally, we provide analysis on the weights and taxonomies induced.

\subsection{Implementation Details}
\label{sec:exp-setting}
\noindent \textbf{Dataset}.
We conduct our experiments on the ImageNet2011 dataset \cite{deng2009imagenet}, which provides a large collection of category items (synsets), with associated images and a label hierarchy (sampled from WordNet) over them. The original ImageNet taxonomy is preprocessed, resulting in a tree structure with 28231 nodes.

\noindent \textbf{Word embedding training}.
We train word embedding for synsets by replacing each word/phrase in a synset with a unique token and then using Google's word2vec tool \cite{mikolov2013distributed}. We combine three public available corpora together, including the latest Wikipedia dump \cite{wikipedia}, the One Billion Word Language Modeling Benchmark \cite{chelba2013one} and the UMBC webbase corpus \cite{han2013umbc}, resulting in a corpus with total 6 billion tokens.
The dimension of the embedding is set to $200$. 

\noindent \textbf{Image processing}.
we employ the ILSVRC12 pre-trained convolutional neural networks \cite{simonyan2014very} to embed each image into the vector space. Then, for each category $x_n$ with images, we estimate a multivariate Gaussian parameterized by $\mathcal{N}_{x_n} = (\mu_{x_n}, \Sigma_{x_n})$, and constrain $\Sigma_{x_n}$ to be diagonal to prevent overfitting. For categories with very few images, we only estimate a mean vector $\mu_{x_n}$. For nodes that do not have images, we ignore the visual feature.

\noindent \textbf{Training configuration}.
The feature vector is a concatenation of 6 parts, as detailed in section \ref{sec:feat}. All pairwise distances are precomputed and stored in memory to accelerate Gibbs sampling. The initial learning rate for gradient descent in the M step is set to 0.1, and is decreased by a fraction of 10 every 100 EM iterations.

\subsection{Evaluation}
\subsubsection{Experimental Settings}
\label{sec:exp-settings}
We evaluate our model on three subtrees sampled from the ImageNet taxonomy. To collect the subtrees, we start from a given root (e.g. consumer goods) and traverse the full taxonomy using BFS, and collect all descendant nodes within a depth $h$ (number of nodes in the longest path). We vary $h$ to get a series of subtrees with increasing heights $h \in \{4, 5, 6, 7\}$ and various scales (maximally 1326 nodes) in different domains. The statistics of the evaluation sets are provided in Table \ref{tab:statistics}. To avoid ambiguity, all nodes used in ILSVRC 2012 are removed as the CNN feature extractor is trained on them.

\begin{table}[t]
\small
\centering
    \begin{tabular}{| c | c | c | c |}
    \hline
    Trees & Tree A & Tree B & Tree C \\
    \hline \hline
    \textbf{Synset ID} &  12638 & 19919  & 23733 \\ \hline
    \textbf{Name} &  consumer goods &  animal & food, nutrient \\ \hline
    \hline
    $h=4$ &  $187$ & $207$ & $572$ \\ \hline
    $h=5$ &  $362$ & $415$  & $890$ \\ \hline
    $h=6$ &  $493$ & $800$ & $1166$ \\ \hline
	$h=7$ &  $524$ & $1386$ & $1326$ \\ \hline
    \end{tabular}
\caption{Statistics of our evaluation set. The bottom 4 rows give the number of nodes within each height $h \in \{4,5,6,7\}$. The scale of the threes range from small to large, and there is no overlapping among them.}
\label{tab:statistics}
\end{table}

We design two different tasks to evaluate our model. (1) In the \emph{hierarchy completion} task, we randomly remove some nodes from a tree and use the remaining hierarchy for training. In the test phase, we infer the parent of each removed node and compare it with groundtruth. This task is designed to figure out whether our model can successfully induce hierarchical relations after learning from within-domain parent-child pairs. (2) Different from the previous one, the \emph{hierarchy construction} task is designed to test the generalization ability of our model, i.e. whether our model can learn statistical patterns from one hierarchy and transfer the knowledge to build a taxonomy for another collection of out-of-domain labels. Specifically, we select two trees as the training set to learn $\bm{w}$. In the test phase, the model is required to build the full taxonomy from scratch for the third tree.

We use \emph{Ancestor $F_1$} as our evaluation metric \cite{kozareva2010semi,navigli2011graph,bansalstructured}. Specifically, we measure $F_1 = 2 P R / (P + R)$ values of predicted ``is-a'' relations
where the precision (P) and recall (R) are:
\begin{equation}
\small
\begin{aligned}
& P = \frac{|\mbox{isa}_{\mbox{predicted}} \cap \mbox{isa}_{\mbox{gold}}|} {|\mbox{isa}_{\mbox{predicted}}|},
R = \frac{|\mbox{isa}_{\mbox{predicted}} \cap \mbox{isa}_{\mbox{gold}}|} {|\mbox{isa}_{\mbox{gold}}|}. \nonumber
\end{aligned}
\end{equation}

We compare our method to two previously state-of-the-art models by \newcite{fu2014learning} and \newcite{bansalstructured}, which are closest to ours.

\subsubsection{Results}
\label{sec:exp-external}
\noindent \textbf{Hierarchy completion.}
In the \emph{hierarchy completion} task, we split each tree into $70\%$ nodes for training and $30\%$ for test, and experiment with different $h$. We compare the following three systems: (1) \textit{Fu2014}\footnote{We tried different parameter settings for the number of clusters $C$ and the identification threshold $\delta$, and reported the best performance we achieved.} \cite{fu2014learning}; (2) \textit{Ours (L)}: Our model with only language features enabled (i.e. surface features, parent-child word-word relation feature and siblings word-word relation feature); (3) \textit{Ours (LV)}: Our model with both language features and visual features \footnote{In the comparisons to \cite{fu2014learning} and \cite{bansalstructured}, we simply set $K = \infty$, \ie we use all available images of the parent category to estimate the PC-V1 feature.}.
The average performance on three trees are reported at Table \ref{tab:comparison}. We observe that the performance gradually drops when $h$ increases, as more nodes are inserted when the tree grows higher, leading to a more complex and difficult taxonomy to be accurately constructed. Overall, our model outperforms Fu2014 in terms of the $F_1$ score, even without visual features. In the most difficult case with $h = 7$, our model still holds an $F_1$ score of 0.42 ($2 \times$ of Fu2014), demonstrating the superiority of our model.

\begin{table}[tbp]
\small
  \centering
    \begin{tabular}{  |l | c | c | c | c| }
    \hline
    Method & $h=4$ & $h=5$ & $h=6$  & $h=7$ \\ \hline
    \hline
    \multicolumn{5}{|c|}{Hierarchy Completion}\\ \hline
    Fu2014 & 0.66 & 0.42 & 0.26 & 0.21  \\ \hline
    Ours (L) & 0.70  & 0.49 & 0.45 & 0.37\\ \hline
    Ours (LV)& \textbf{0.73} & \textbf{0.51} & \textbf{0.50} & \textbf{0.42} \\ \hline
    \hline
    \multicolumn{5}{|c|}{Hierarchy Construction}\\ \hline
    Fu2014 & 0.53 & 0.33 & 0.28 & 0.18 \\ \hline
    Bansal2014 & 0.67 & 0.53 & 0.43 & 0.37 \\ \hline
    Ours (L) & 0.58 & 0.41 & 0.36 & 0.30 \\ \hline
    Ours (LB) & 0.68 & 0.55  & 0.45 & 0.40\\ \hline
    Ours (LV) & 0.66 & 0.52 & 0.42 & 0.34 \\ \hline
    Ours (LVB - E) & 0.68 & 0.55 & 0.44 & 0.39 \\ \hline
    Ours (LVB) & \textbf{0.70} & \textbf{0.57} & \textbf{0.49} & \textbf{0.43}\\
    \hline
    \end{tabular}
    \vspace{-5pt}
    \caption{Comparisons among different variants of our model, \newcite{fu2014learning} and \newcite{bansalstructured} on two tasks. The ancestor-$F_1$ scores are reported.}
    \vspace{-10pt}
    \label{tab:comparison}
\end{table}

\noindent \textbf{Hierarchy construction.}
The hierarchy construction task is much more difficult than hierarchy completion task because we need to build a taxonomy from scratch given only a hyper-root.
For this task, we use a leave-one-out strategy, i.e. we train our model on every two trees and test on the third, and report the average performance in Table \ref{tab:comparison}.
We compare the following methods: (1) \textit{Fu2014}, (2) \textit{Ours (L)}, and (3) \textit{Ours (LV)}, as described above;
(4) \textit{Bansal2014}: The model by \newcite{bansalstructured} retrained using our dataset;
(5) \textit{Ours (LB)}: By excluding visual features, but including other language features from \newcite{bansalstructured};
(6) \textit{Ours (LVB)}: Our full model further enhanced with all semantic features from \newcite{bansalstructured}; (7) \textit{Ours (LVB - E)}: By excluding word embedding-based language features from \textit{Ours (LVB)}.

As shown, on the hierarchy construction task, our model with only language features still outperforms Fu2014 with a large gap ($0.30$ compared to $0.18$ when $h=7$), which uses similar embedding-based features. The potential reasons are two-fold. First, we take into account not only parent-child relations but also siblings. Second, their method is designed to induce only pairwise relations. To build the full taxonomy, they first identify all possible pairwise relations using a simple thresholding strategy and then eliminate conflicted relations to obtain a legitimate tree hierarchy. In contrast, our model is optimized over the full space of all legitimate taxonomies by taking the \emph{structure operation} in account during Gibbs sampling.

When comparing to Bansal2014, our model with only word embedding-based features underperforms theirs. However, when introducing visual features, our performance is comparable (p-value = 0.058).
Furthermore, if we discard visual features but add semantic features from \newcite{bansalstructured}, we achieve a slight improvement of 0.02 over Bansal2014 (p-value = 0.016), which is largely attributed to the incorporation of word embedding-based features that encode high-level linguistic regularity.
Finally, if we enhance our full model with all semantic features from \newcite{bansalstructured}, our model outperforms theirs by a gap of 0.04 (p-value $<$ 0.01), which justifies our intuition that perceptual semantics underneath visual contents are quite helpful.

\subsection{Qualitative Analysis}
\label{sec:exp-qualitative}
In this section, we conduct qualitative studies to investigate \emph{how} and \emph{when} the visual information helps the taxonomy induction task.

\noindent \textbf{Contributions of visual features.}
To evaluate the contribution of each part of the visual features to the final performance, we train our model jointly with textual features and different combinations of visual features, and report the ancestor-$F_1$ scores. As shown in Table \ref{tab:feature_comp}. When incorporating the feature S-V1, the performance is substantially boosted by a large gap at all heights, showing that visual similarity between sibling nodes is a strong evidence for taxonomy induction. It is intuitively plausible, as it is highly likely that two specific categories share a common (and more general) parent category if similar visual contents are observed between them. Further, adding the PC-V1 feature gains us a better improvement than adding PC-V2, but both minor than S-V1.

\begin{figure}[t]
\small
  \centering
    \includegraphics[width=0.48\textwidth]{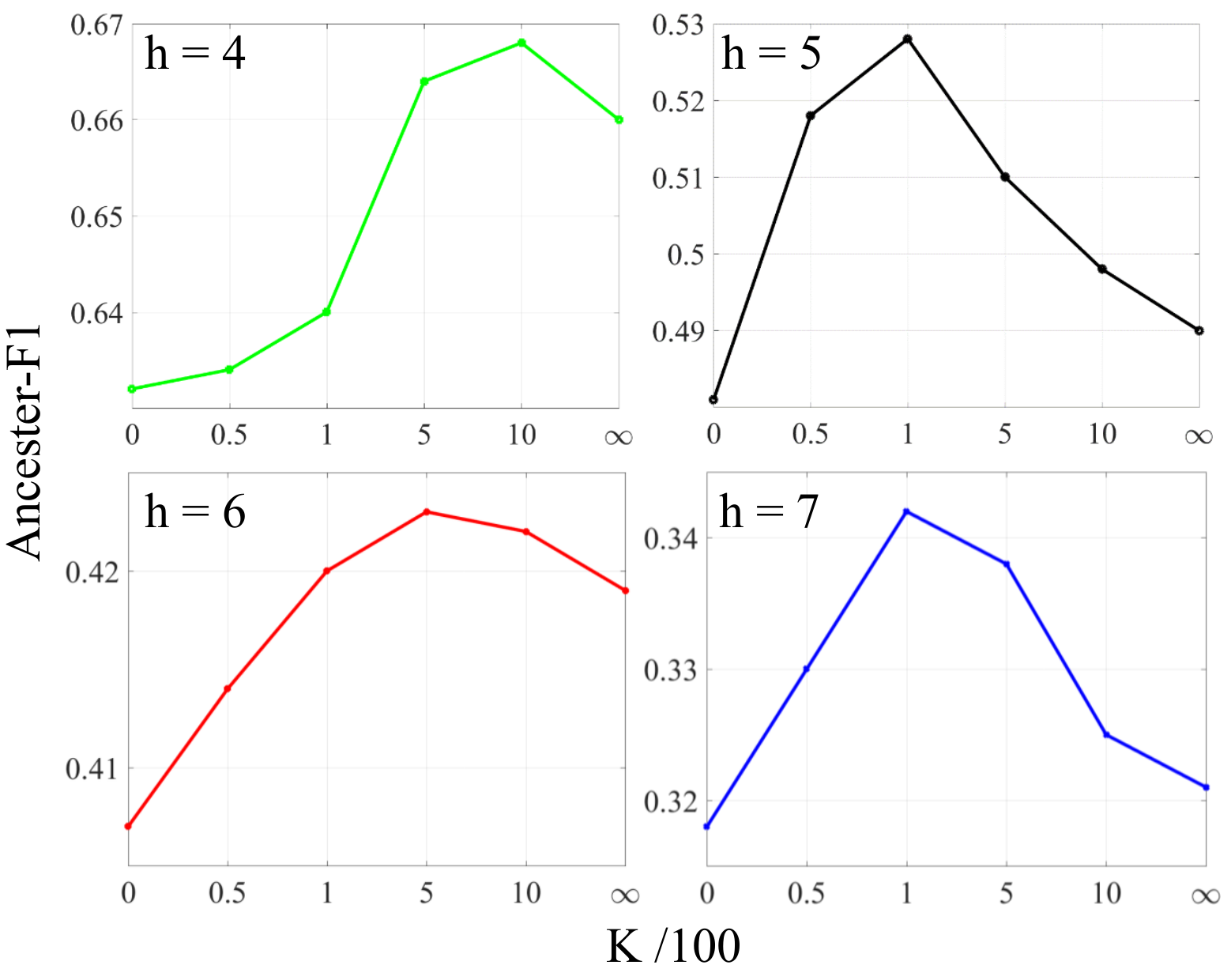}
    \vspace{-10pt}
    \caption{The Ancestor-$F_1$ scores changes over K (number of images used in the PC-V1 feature) at different heights. The values in the x-axis are $K/100$; $K = \infty$ means all images are used.}
    \label{fig:k}
\end{figure}

Compared to that of siblings, the visual similarity between parents and children does not strongly holds all the time. For example, images of \emph{Terrestrial animal} are only partially similar to those of \emph{Feline}, because the former one contains the later one as a subset. Our feature captures this type of ``contain'' relation between parents and children by considering only the top-$K$ images from the parent category that have highest probabilities under the Gaussian distribution of the child category. To see this, we vary $K$ while keep all other settings, and plot the $F_1$ scores in Fig \ref{fig:k}. We observe a trend that when we gradually increase $K$, the performance goes up until reaching some maximal; It then slightly drops (or oscillates) even when more images are available, which confirms with our feature design that only top images should be considered in parent-child visual similarity.

Overall, the three visual features complement each other, and achieve the highest performance when combined.

\begin{table}[t]
\small
    \centering
    \begin{tabular}{ | c c c | c | c | c | c |}
    \hline
    S-V1 & PC-V1 & PC-V2 & h = 4 & h = 5 & h = 6 & h = 7 \\
    \hline \hline
      & & & 0.58 & 0.41 & 0.36 & 0.30\\ \hline
      \checkmark & & & 0.63 & 0.48 & 0.40 & 0.32 \\ \hline
       & \checkmark & & 0.61 & 0.44 & 0.38 & 0.31 \\ \hline
       & & \checkmark & 0.60 & 0.42 & 0.37 & 0.31 \\ \hline
      \checkmark & \checkmark & & 0.65 & \textbf{0.52} & 0.41 & 0.33 \\ \hline
      \checkmark & \checkmark & \checkmark & \textbf{0.66} & \textbf{0.52} & \textbf{0.42} & \textbf{0.34} \\ \hline
    \end{tabular}
    \vspace{-5pt}
\caption{The performance when different combinations of visual features are enabled.}
\vspace{-8pt}
\label{tab:feature_comp}
\end{table}

\begin{figure}[tbp]
\small
  \centering
    \includegraphics[width=0.5\textwidth]{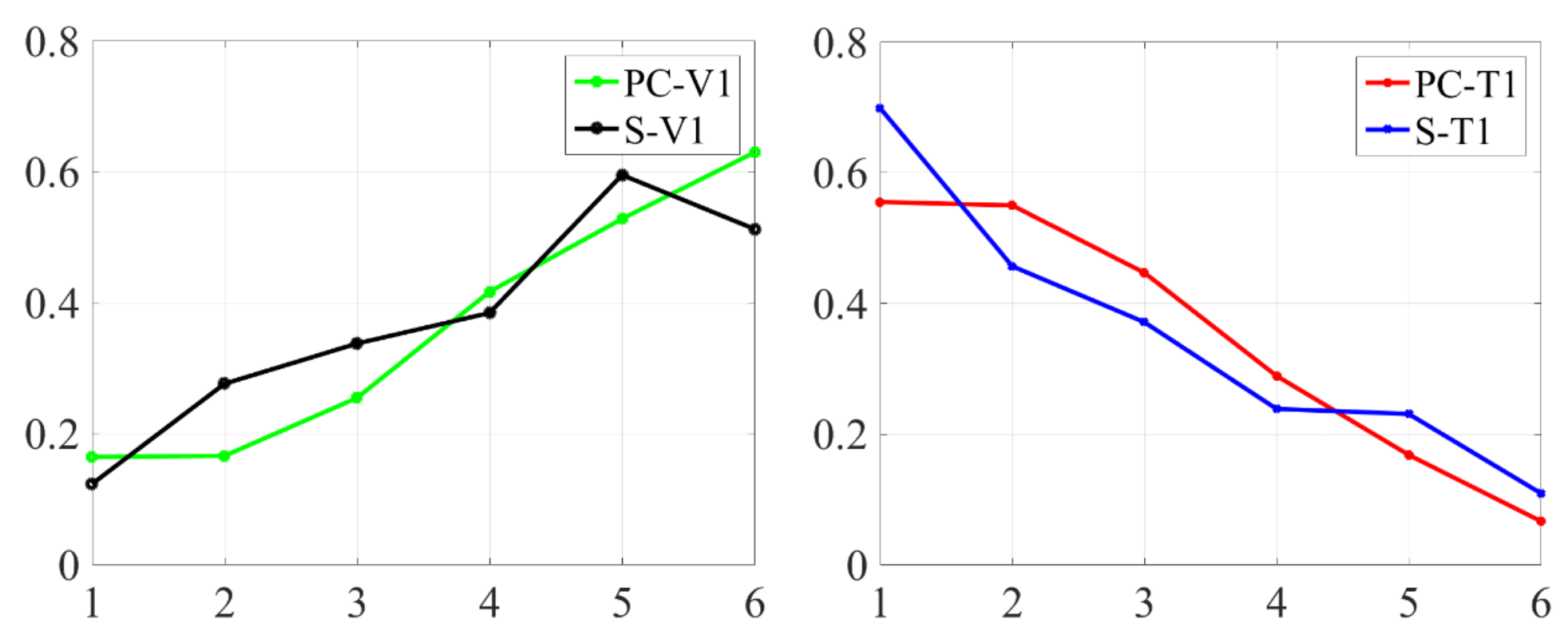}
    \vspace{-10pt}
    \caption{Normalized weights of each feature v.s. the layer depth.}
  \label{fig:hierarchy}
\end{figure}

\noindent \textbf{Visual representations.} To investigate how the image representations affect the final performance, we compare the ancestor-F1 score when different pre-trained CNNs are used for visual feature extraction. Specifically, we employ both the CNN-128 model (128 dimensional feature with $15.6\%$ top-5 error on ILSVRC12) and the VGG-16 model (4096 dimensional feature with $7.5\%$ top-5 error) by \newcite{simonyan2014very}, but only observe a slight improvement of 0.01 on the ancestor-F1 score for the later one.

\noindent \textbf{Relevance of textual and visual features v.s. depth of tree.}
Compared to \newcite{bansalstructured}, a major difference of our model is that different layers of the taxonomy correspond to different weights $\bm{w}_l$, while in \cite{bansalstructured} all layers share the same weights. Intuitively, introducing layer-wise $\bm{w}$ not only extends the model capacity, but also differentiates the importance of each feature at different layers.
For example, the images of two specific categories, such as \emph{shark} and \emph{ray}, are very likely to be visually similar. However, when the taxonomy goes from bottom to up (specific to general), the visual similarity is gradually undermined ---
images of \emph{fish} and \emph{terrestrial animal} are not necessarily similar any more. Hence, it is necessary to privatize the weights $\bm{w}$ for different layers to capture such variations, i.e. the visual features become more and more evident from shallow to deep layers, while the textual counterparts, which capture more abstract concepts, relatively grow more indicative oppositely from specific to general.

\begin{figure*}[t]
\centering
\includegraphics[width=16cm]{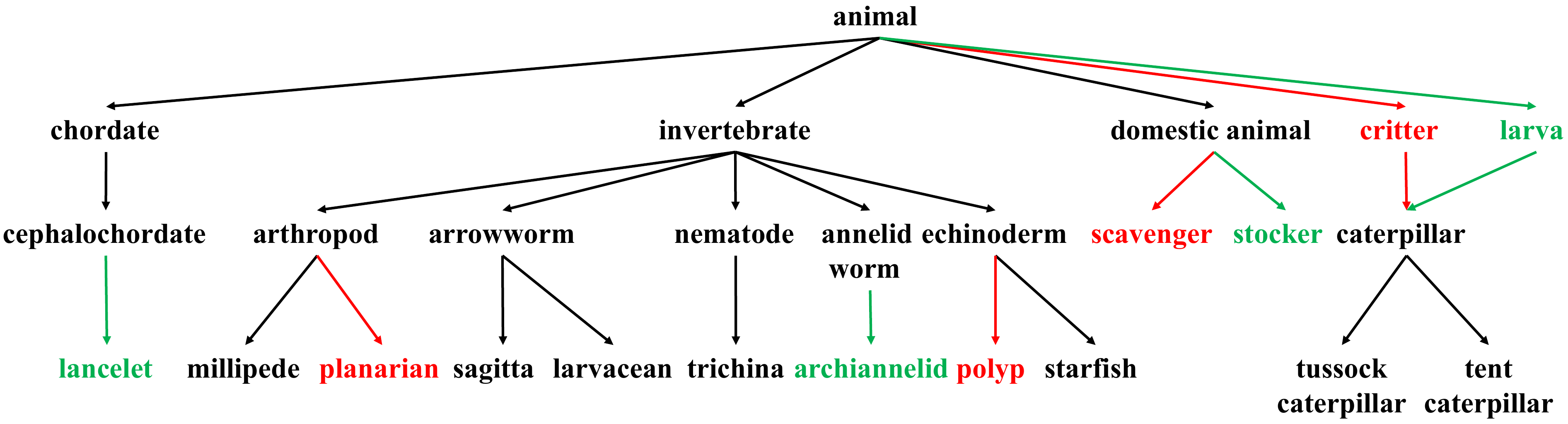}
\includegraphics[width=16cm]{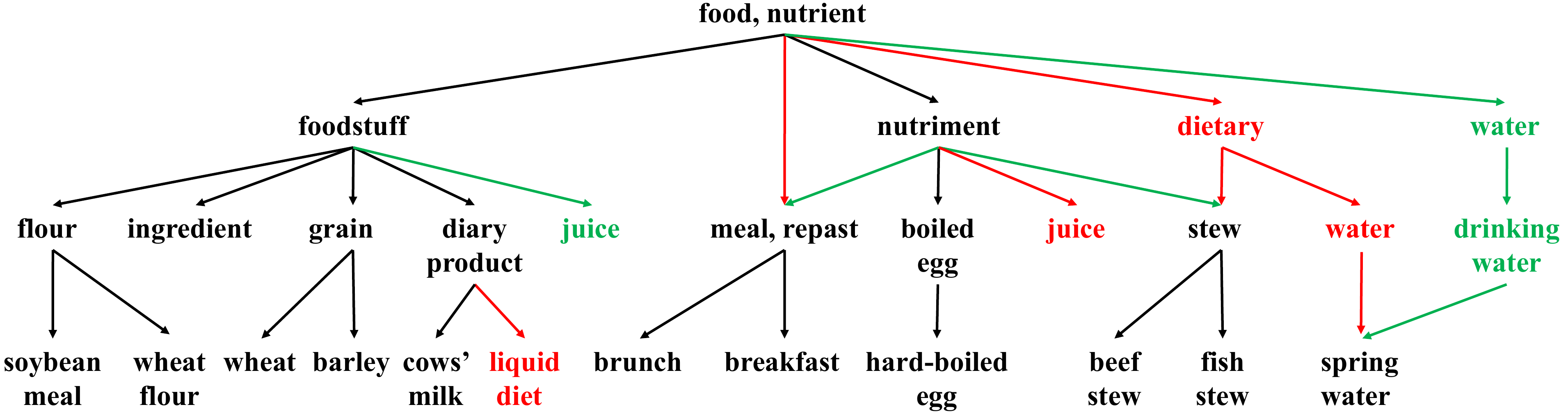}
\vspace{-15pt}
\caption{Excerpts of the prediction taxonomies, compared to the groundturth. Edges marked as red and green are false predictions and unpredicted groundtruth links, respectively.}
\vspace{-5pt}
\label{fig:excerpt1}
\end{figure*}

To visualize the variations across layers, for each feature component, we fetch its corresponding block in $\bm{w}$ as $V$. Then, we average $|V|$ and observe how its values change with the layer depth $h$. For example, for the parent-child word-word relation feature, we first fetch its corresponding weights $V$ from $\bm{w}$ as a $20\times 6$ matrix, where 20 is the feature dimension and $6$ is the number of layers. We then average its absolute values\footnote{We take the absolute value because we only care about the relevance of the feature as an evidence for taxonomy induction, but note that the weight can either encourage (positive) or discourage (negative) connections of two nodes.} in column and get a vector $v$ with length 6. After $\ell_2$ normalization, the magnitude of each entry in $v$ directly reflects the relative importance of the feature as an evidence for taxonomy induction. Fig \ref{fig:hierarchy}(b) plots how their magnitudes change with $h$ for every feature component averaged on three train/test splits.
It is noticeable that for both word-word relations (S-T1, PC-T1), their corresponding weights slightly decrease as $h$ increases. On the contrary, the image-image relation features (S-V1, PC-V1) grows relatively more prominent. The results verify our conjecture that when the category hierarchy goes deeper into more specific classes, the visual similarity becomes relatively more indicative as an evidence for taxonomy induction.

\noindent \textbf{Visualizing results.}
Finally, we visualize some excerpts of our predicted taxonomies, as compared to the groundtruth in Fig \ref{fig:excerpt1}.

\section{Conclusion}
In this paper, we study the problem of automatically inducing semantically meaningful concept taxonomies from multi-modal data. We propose a probabilistic Bayesian model which leverages distributed representations for images and words. We compare our model and features to previous ones on two different tasks using the ImageNet hierarchies, and demonstrate superior performance of our model, and the effectiveness of exploiting visual contents for taxonomy induction. We further conduct qualitative studies and distinguish the relative importance of visual and textual features in constructing various parts of a taxonomy.

\section*{Acknowledgements}
We would like to thank anonymous reviewers for their valuable feedback. We would also like to thank Mohit Bansal for helpful suggestions. We thank NVIDIA for GPU donations. The work is supported by NSF Big Data IIS1447676.

\bibliography{hierarchy}
\bibliographystyle{acl2016}

\end{document}